\documentclass[runningheads]{llncs}

\usepackage[T1]{fontenc}
\usepackage{graphicx}
\usepackage{amsmath,amssymb}
\usepackage{booktabs}
\usepackage{hyperref}

\begin{document}

\title{Evaluating Explainability in Safety-Critical ATR Systems:
Limitations of Post-Hoc Methods and Paths Toward Robust XAI}

\titlerunning{Evaluating XAI in Safety-Critical ATR}

\author{
Vanessa Buhrmester\inst{1}\orcidID{0000-0003-2356-1313} \and
David Muench\inst{1}\orcidID{0000-0002-8577-5256} \and
Dimitri Bulatov\inst{1}\orcidID{0000-0002-0560-2591} \and
Michael Arens\inst{1}\orcidID{0000-0002-8577-5256}
}

\authorrunning{V. Buhrmester et al.}

\institute{
Fraunhofer Institute of Optronics, System Technologies and Image Exploitation IOSB, Ettlingen, Germany\\
\email{vanessa.buhrmester@iosb.fraunhofer.de, david.muench@iosb.fraunhofer.de}
}

\maketitle

\begin{abstract}
Explainable Artificial Intelligence (XAI) is increasingly recognized as essential for deploying machine learning systems in safety-critical environments. In Automatic Target Recognition (ATR), where models operate on image, video, radar, and multisensor data, high predictive performance alone is insufficient. Model decisions must also be interpretable, reliable, and suitable for validation.

This paper presents a structured evaluation of explainability methods in the context of safety-critical ATR systems: We identify major XAI paradigms, including saliency-based, attention-based, and surrogate approaches, as well as recent detection-aware extensions. 
Based on this, we formalize explainability as an assurance-oriented assessment problem, introduce a taxonomy, and assess these methods with respect to four key dimensions: interpretability, robustness, vulnerability to manipulation, and suitability for validation and verification.
The analysis identifies systematic limitations of current post-hoc explanation methods. In particular, we derive critical failure modes such as spurious explanations, instability under perturbations, and overtrust induced by visually convincing outputs. These findings indicate that widely used XAI techniques may be insufficient for safety-critical deployment.

Finally, we discuss implications for ATR systems and outline directions toward more robust, causally grounded, and physically informed explainability methods. Our results emphasize the need to move beyond visually plausible explanations toward approaches that support reliable decision-making and system-level assurance.
\keywords{Explainable AI \and Automatic Target Recognition \and Object Detection \and Safety-Critical AI \and Evaluation}
\end{abstract}

\section{Introduction}

Explainable Artificial Intelligence (XAI) is increasingly regarded as a prerequisite for deploying machine learning systems in safety-critical environments. This is particularly evident in Automatic Target Recognition (ATR), where models operate on heterogeneous data sources such as imagery, video streams, radar, and multisensor data. In these settings, predictive accuracy alone is insufficient. Model decisions must also be transparent, reliable, and suitable for technical assessment, since misinterpretations may have significant operational consequences.

Despite their strong performance, modern deep learning models, especially convolutional and transformer-based architectures, remain inherently opaque. A broad range of post-hoc explanation techniques has been proposed to address this issue, including saliency-based, attention-based, and surrogate approaches \cite{buhrmester2021analysis,lundberg2017shap,ribeiro2016lime,selvaraju2017gradcam}. While these methods often produce visually convincing explanations, their reliability is increasingly questioned. Empirical studies indicate that explanations can remain stable under model randomization, react strongly to minor perturbations, or fail to reflect the underlying decision process altogether \cite{adebayo2018sanity}.

These limitations are particularly critical in the ATR context. Unlike standard image classification tasks, ATR systems must operate under varying environmental conditions, sensor noise, incomplete observations, and real-time constraints. Furthermore, explanations are not only used for human interpretation, but also for system validation, robustness analysis, and decision support. Consequently, explanation methods must satisfy stronger requirements, including stability, resistance to manipulation, and compatibility with validation and verification processes.

Recent research has emphasized the need for structured evaluation of XAI methods, especially in object detection scenarios where explanations must capture both spatial localization and semantic relevance. At the same time, application-driven studies in domains such as remote sensing and UAV-based perception highlight the growing importance of explainability in real-world and safety-critical settings.

This work provides a structured analysis of XAI methods in the context of modern ATR systems, focusing on saliency-based, attention-based, and surrogate-based approaches, including recent detection-aware extensions, and analyzing their strengths, limitations, and common failure modes in safety-critical settings. This paper does not propose a new explanation algorithm. Instead, it provides an ATR-specific assessment framework for evaluating whether existing XAI paradigms are suitable for safety-critical use, with particular emphasis on robustness, manipulation resistance, and validation and verification.

The main contributions of this work are as follows:
\begin{itemize}
\item We formalize explainability in safety-critical ATR as an assurance-oriented assessment problem and introduce four evaluation dimensions: interpretability, robustness, vulnerability to manipulation, and suitability for validation and verification.
\item We provide an ATR-specific, unified taxonomy of XAI methods, covering saliency-based, attention-based, surrogate, detection-aware, concept-based, and intrinsic or physics-informed approaches.
\item We present a systematic, cross-paradigm evaluation of major XAI methods with respect to these dimensions, identifying their suitability for exploratory analysis, structured validation, and high-assurance deployment.
\item We identify systematic gaps between current post-hoc XAI practice and safety-critical requirements, including instability, spurious explanations, overtrust, and limited integration into V\&V workflows.
\item We derive concrete research directions toward robust, causally grounded, and physically informed XAI methods, highlighting the need for explanations that support verification, resist manipulation, and align with domain constraints.
\end{itemize}

\section{Related Work}

\subsection{XAI in Computer Vision}

Explainable Artificial Intelligence has been widely studied in computer vision, where most approaches can be grouped into attribution-based, activation-based, perturbation-based, concept-based, and transformer-oriented methods. Recent survey papers provide increasingly structured overviews of this landscape and highlight the growing shift from image classification toward more complex tasks such as object detection and tracking \cite{cheng2025comprehensive,mi2024toward}. Attribution-based methods such as gradient saliency and Integrated Gradients  \cite{sundararajan2017integrated} focus on local feature relevance, while activation-based approaches such as CAM and Grad-CAM exploit intermediate feature maps for spatial localization. More recent reviews also emphasize transformer-based explainability methods and discuss their relevance for modern vision architectures \cite{cheng2025comprehensive}.

At the same time, a growing body of work points to fundamental shortcomings of existing XAI approaches. Visually coherent explanations do not necessarily correspond to model-relevant features, and many methods exhibit pronounced sensitivity to noise and perturbations. These observations have motivated a shift toward more critical evaluation protocols and application-specific analysis frameworks, especially in high-stakes domains.

\subsection{Explainability for Object Detection}

Compared with image classification, explainability in object detection poses additional challenges because explanations must account not only for class predictions but also for localization decisions and multi-object scenarios. Classical saliency-based approaches such as Grad-CAM \cite{selvaraju2017gradcam} have therefore been extended to object detection pipelines such as Faster R-CNN \cite{ren2015faster} and YOLO \cite{redmon2016yolo}, where explanations are typically conditioned on specific detection heads or bounding boxes.

Recent work has proposed detection-aware extensions of CAM-based methods, including the Gaussian-Class Activation Mapping Explainer (G-CAME) \cite{gcame2024}, which improves instance-level localization and computational efficiency in object detection settings. In parallel, there is growing interest in benchmarking explainability methods specifically for detection tasks. ODExAI, for example, introduces a dedicated evaluation framework for object detection explainability based on localization accuracy, faithfulness, and computational complexity \cite{nguyen2025odexai}. Such work is particularly relevant for ATR scenarios, where explanations must capture both semantic target evidence and spatial precision.

Transformer-based object detectors such as DETR \cite{carion2020detr} have further expanded the methodological landscape. Their attention mechanisms are often interpreted as explanatory signals, but prior work has shown that attention weights do not necessarily correspond to causal relevance \cite{jain2019attention}. This limits their direct interpretability and calls for critical assessment in practical use cases.

\subsection{XAI in Safety-Critical and ATR Applications}
\begin{figure}
\centering
\includegraphics[width=1\linewidth]{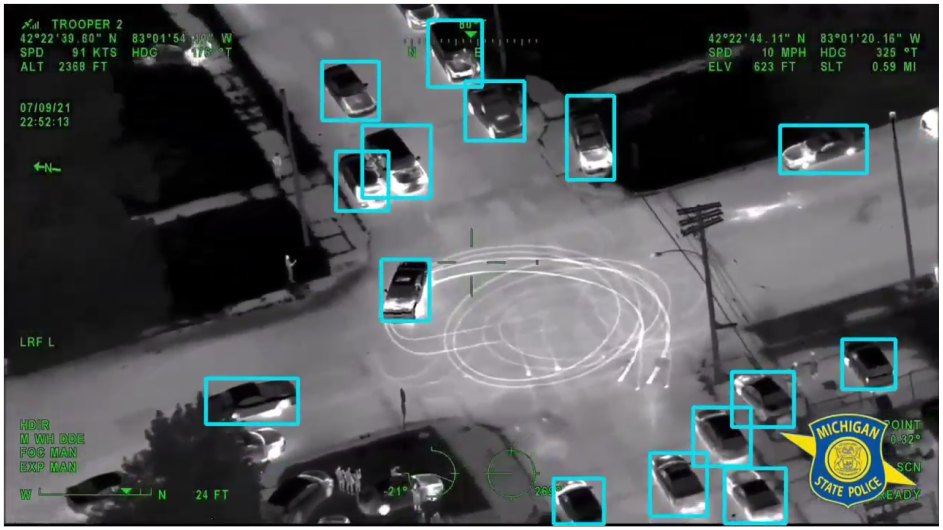}
\caption{\label{fig:IR} Illustrative example of safety-critical applications: Real-time automatic target recognition from a UAV-based thermal infrared perspective}
\end{figure}

The use of XAI in safety-critical applications has received increasing attention, particularly in domains where incorrect or poorly understood model decisions may lead to severe consequences. In such contexts, explainability is not only a diagnostic tool for model developers, but also a prerequisite for trust, verification, and operational acceptance.

In the ATR domain, explainability has been studied across multiple sensing modalities, including image, video, radar, and SAR data. Image-based ATR and object detection systems predominantly rely on post-hoc saliency or attention-based approaches, while radar- and SAR-based ATR increasingly explore more structured or physically grounded forms of interpretability. Application-driven research in satellite imagery demonstrates that explainable object detection can improve robustness and support structured reasoning in remote sensing scenarios \cite{roy2025satellite}. Similarly, explainability is becoming increasingly relevant in monocular vision-based UAV systems, where transparent obstacle detection and navigation decisions are critical for safe and trustworthy autonomous operation \cite{javaid2025uav}.

Overall, prior work shows that explainability in ATR is moving beyond generic visualization methods toward task-specific, detection-aware, and application-driven approaches. However, the literature still lacks a unified evaluation perspective tailored to safety-critical ATR systems. In particular, there remains a need for systematic comparison of XAI methods with respect to robustness, manipulability, and suitability for validation and verification.

\subsection{Gap Analysis and Positioning}
Although recent surveys provide comprehensive overviews of XAI methods in computer vision, they remain largely application-agnostic and do not fully address the requirements of safety-critical ATR systems. Similarly, detection-aware XAI frameworks improve the evaluation of explanations for object detection, but they primarily focus on localization, faithfulness, and computational aspects. These criteria are necessary but insufficient for high-assurance ATR applications, where explanations must also be robust under sensor and environmental perturbations, resistant to manipulation, and usable within validation and verification processes.
This paper therefore positions itself between general XAI surveys, object-detection-specific explanation benchmarks, and safety-critical AI research. Its contribution is an ATR-specific assessment framework that links major XAI paradigms to operational assurance requirements. The goal is not to introduce a new explanation method, but to identify which classes of explanations are suitable, insufficient, or promising for safety-critical ATR.

\par\vspace{\baselineskip}
The discussion above shows that existing XAI literature provides either broad methodological taxonomies, object-detection-specific explanation techniques, or general critiques of post-hoc explanations in high-stakes domains. What is still missing is an ATR-specific perspective that connects explanation methods to operational assurance requirements. The following section therefore formulates XAI evaluation in ATR as a multi-criteria assessment problem.

\section{Evaluation Framework}

\subsection{Problem Formulation}

Automatic Target Recognition systems can be formalized as predictive models that map sensor data to target labels. Given an input space $\mathcal{X}\subseteq \mathbb{R}^n$ and output space $\mathcal{Y}$, e.g., continuous predictions. An input $x \in \mathcal{X}$, representing data from one or multiple sensors, a model $f$ produces a prediction $y \in \mathcal{Y}$, typically by maximizing the posterior probability:
\begin{equation}
f(x) = \arg\max_{y \in \mathcal{Y}} p(y \mid x).
\end{equation}

In modern ATR systems, $f$ is commonly implemented as a deep neural network, such as a convolutional or transformer-based architecture. In object detection settings, the model output may additionally include spatial information, such as bounding boxes or segmentation masks, resulting in structured predictions of the form:
\begin{equation}
f(x) = \{(y_i, b_i)\}_{i=1}^{N},
\end{equation}
where $y_i$ denotes the predicted class, $b_i$ the corresponding spatial localization as bounding box and $N$ the number of detections.

To improve transparency, an explanation function $E$ is introduced, which maps a model $f$ and an input $x$ to an interpretable representation:
\begin{equation}
E: (f,x) \mapsto e,
\end{equation}
where the explanation $e$ may take different forms, such as saliency maps, attention distributions, or surrogate model approximations.

The central objective of explainability is to ensure that $e$ provides meaningful insight into the decision-making process of $f$. In safety-critical ATR systems, however, this objective is subject to additional requirements. Explanations must satisfy properties that go beyond visual interpretability, including stability under input perturbations, resistance to manipulation, and consistency with the underlying model behavior.

This leads to a fundamental challenge: given a model $f$ and an explanation method $E$, how can the quality of the resulting explanation $e$ be assessed in a principled and application-relevant manner? In contrast to standard machine learning evaluation, there is typically no ground truth for explanations \cite{kadir2023evaluation,lakkaraju2023m4}, which makes it difficult to directly measure correctness.

ATR systems introduce additional complexity due to multisensor inputs, varying environmental conditions, and real-time constraints. Explanations must therefore capture both semantic relevance, i.e., which features influence the predicted class, and spatial localization, i.e., where relevant information is located, while remaining robust across different sensing modalities.

In this work, we address this challenge by formulating the evaluation of explainability methods as a multi-criteria problem. Specifically, we assess explanation methods along four key dimensions: interpretability, robustness, vulnerability to manipulation, and suitability for validation and verification. This formulation provides the basis for the comparative analysis presented in the following sections.

\subsection{Evaluation Criteria}
\label{sec: Evaluation Criteria}

Based on the problem formulation introduced above, the evaluation of explainability methods in ATR systems can be formulated as a multi-dimensional assessment problem. Since no ground truth explanations are available in most practical scenarios, explanation quality must be evaluated indirectly through a set of well-defined criteria.

We consider four dimensions that are particularly relevant for safety-critical ATR applications: interpretability, robustness, vulnerability to manipulation, and suitability for validation and verification.

\textbf{Interpretability} refers to the extent to which an explanation can be understood by a human user and provides meaningful insight into the model decision. It is important to distinguish between visual plausibility and semantic correctness. An explanation may appear intuitive while not accurately reflecting the underlying decision process of the model. Therefore, interpretability must be considered together with other criteria.

\textbf{Robustness} describes the stability of an explanation under variations in the input or the model. In practical ATR scenarios, sensor data is often affected by noise, changing viewpoints, or environmental conditions. A robust explanation method should produce consistent outputs under such perturbations. High sensitivity to small input changes reduces explanation reliability and limits practical applicability.

\textbf{Vulnerability to manipulation} captures the susceptibility of explanation methods to adversarial or intentional modifications. Explanations can be altered without significantly affecting the model prediction \cite{laugel2019fooling}, leading to misleading or deceptive interpretations. In safety-critical systems, such behavior poses a significant risk, as it undermines trust in the explanation process.

\textbf{Suitability for validation and verification (V\&V)} refers to the extent to which an explanation method supports systematic analysis of model behavior. This includes the ability to test, validate, and potentially certify model decisions. Explanations that lack structure or consistency are difficult to integrate into formal verification workflows, limiting their usefulness in high-assurance applications.

These criteria are not independent but exhibit inherent trade-offs. For example, highly interpretable explanations may lack robustness, while methods optimized for stability may provide less intuitive representations. In addition, different ATR modalities, such as image-based or radar-based systems, impose different requirements on explanation methods.

The proposed criteria provide a structured basis for comparing XAI methods in the ATR context. In the following section, we apply this framework to analyze the strengths and limitations of different explainability paradigms.

\subsection{Method Categorization}

Based on the evaluation criteria defined above, we categorize explainability methods in the ATR context into five main paradigms: saliency-based methods, attention-based approaches, surrogate models, detection-aware XAI methods, and intrinsic or physics-informed models. This categorization reflects both methodological differences and practical applicability in safety-critical systems.

\textbf{Saliency-based methods} represent the most widely used class of post-hoc explainability techniques. These approaches attribute model predictions to input features by estimating relevance scores, typically derived from gradients or feature activations \cite{simonyan2014saliency,sundararajan2017integrated}. Methods such as Vanilla Gradients, Integrated Gradients, and Grad-CAM \cite{selvaraju2017gradcam} fall into this category. More recent extensions, including detection-aware variants such as G-CAME \cite{gcame2024}, improve spatial localization in object detection scenarios. While saliency methods are easy to apply and broadly compatible with different architectures, they are often limited in terms of robustness and causal interpretability \cite{adebayo2018sanity,ghorbani2019fragile}.

\textbf{Attention-based approaches} leverage internal attention mechanisms, particularly in transformer-based architectures, to provide insight into model behavior \cite{jain2019attention,vaswani2017attention}. In object detection models such as DETR \cite{carion2020detr}, attention weights are often visualized as explanatory signals. These methods offer a global perspective on feature relationships but do not necessarily reflect causal importance. As a result, their interpretability is limited and they should be used with caution in safety-critical applications.

\textbf{Surrogate models and model simplification techniques} aim to approximate complex models with simpler, interpretable representations. Methods such as LIME \cite{ribeiro2016lime} and SHAP \cite{lundberg2017shap} generate local explanations by fitting interpretable models to the behavior of the original model in the vicinity of a given input. While these approaches improve transparency, their reliability depends on approximation fidelity. In high-dimensional ATR scenarios, surrogate models may fail to capture the true decision boundary, leading to misleading explanations.

\textbf{Detection-aware XAI methods} form a distinct paradigm in ATR, as they explicitly incorporate the structure of detection models. These approaches, including detector-specific CAM variants, G-CAME \cite{gcame2024}, and object detection (OD)-specific evaluation methods, align explanations with detection outputs such as bounding boxes or object instances.

In contrast to generic saliency methods, they explicitly couple explanations to object-level predictions, improving spatial localization and interpretability in object detection tasks, which are central to ATR. However, these methods can still inherit limitations from underlying gradient-based techniques, including sensitivity to perturbations and limited causal interpretability.

\textbf{Intrinsic and physics-informed models} integrate interpretability directly into the model architecture. Instead of explaining decisions post hoc, these approaches aim to make the decision process itself transparent. In radar- and SAR-based ATR systems, this often involves incorporating physical models or constraints into the learning process \cite{raissi2019pinn,karniadakis2021piml}. Such methods generally provide higher robustness and stronger alignment with domain knowledge, making them particularly suitable for safety-critical applications.

This categorization highlights fundamental differences between post-hoc and model-inherent explainability approaches. While post-hoc methods offer flexibility and ease of use, intrinsic approaches provide more reliable and structured explanations. The trade-offs between these paradigms are analyzed in the following section using the evaluation criteria defined above.

\section{Comparative Analysis}

\subsection{Comparison of XAI Paradigms}

Using the evaluation criteria introduced in Section~3, we compare the main XAI paradigms with respect to their behavior in ATR scenarios. Rather than providing a purely descriptive overview, the focus lies on identifying systematic strengths and limitations across method classes.

Saliency-based methods provide intuitive and visually accessible explanations by directly mapping relevance to the input space. This makes them particularly attractive for image-based ATR systems, where spatial interpretability is essential. However, their interpretability is often limited to local sensitivity analysis, and their outputs can be highly unstable under small input perturbations. Moreover, these methods lack causal grounding, as highlighted by prior work showing that saliency maps may remain visually similar even when model parameters are randomized \cite{adebayo2018sanity}. This raises concerns about their reliability in safety-critical settings.

Attention-based approaches offer a complementary perspective by capturing global dependencies within the data. In transformer-based architectures, attention weights can provide insight into relationships between different input regions. While this enables a more holistic view of model behavior, attention mechanisms do not necessarily reflect causal importance. Empirical studies have demonstrated that significantly different attention distributions can lead to similar model outputs, limiting their interpretability and robustness.

Surrogate models aim to improve interpretability by approximating complex models with simpler, more transparent representations. This enables structured explanations and facilitates human understanding of decision boundaries. However, approximation fidelity is critical. In high-dimensional ATR scenarios, surrogate models may fail to accurately capture the behavior of the original model, resulting in misleading explanations. Their robustness is therefore dependent on the stability of the approximation process.

In addition to these general post-hoc approaches, detection-aware XAI methods bridge the gap between generic saliency approaches and task-specific requirements of object detection. By aligning explanations with detection outputs such as bounding boxes or object instances, they improve spatial interpretability in ATR scenarios. However, they still rely on underlying gradient-based mechanisms and therefore inherit key limitations in robustness and causal validity.

In contrast, intrinsic and physics-informed models integrate interpretability directly into the model architecture. These approaches often leverage domain knowledge, such as physical signal properties in radar or SAR systems, to produce explanations that are both meaningful and consistent with underlying processes. As a result, they tend to exhibit higher robustness and lower susceptibility to manipulation. However, this comes at the cost of reduced flexibility and increased modeling complexity.\\
Overall, the comparison reveals fundamental trade-offs between interpretability, robustness, and methodological complexity. Post-hoc methods such as saliency and attention provide flexible and easily deployable explanations but suffer from limitations in stability and causal validity. Intrinsic approaches, while more reliable, require stronger assumptions and domain-specific modeling.

\begin{table}[htp!]
\centering
\caption{Assessment of XAI paradigms according to safety-critical ATR requirements.}
\label{tab:xai_assessment_atr}
\small
\begin{tabular}{p{1.8cm} p{1.8cm} p{1.9cm} p{2cm} p{1.8cm} p{2.5cm}}
\hline
\textbf{XAI paradigm} &
\textbf{Interpreta\-bility} &
\textbf{Robustness} &
\textbf{Vulnerability to manipulation} &
\textbf{V\&V suitability} &
\textbf{ATR-specific suitability} \\
\hline

Saliency-based methods, e.g., Vanilla Gradients, Integrated Gradients, Grad-CAM &
High visual accessibility; intuitive spatial overlays for image-based ATR &
Often low; sensitive to noise, small perturbations, preprocessing, and model changes &
High; explanations can be altered or remain visually plausible despite weak relation to the model &
Limited; difficult to use as reliable evidence due to instability and lack of causal grounding &
Useful for initial inspection of EO/IR image-based ATR and object detection, but insufficient as standalone evidence in safety-critical settings \\

\hline

Attention-based methods &
Medium; can indicate global relations between image regions or tokens &
Medium to low; attention patterns may vary and do not necessarily reflect causal relevance &
Medium; attention can be misleading if interpreted as explanation without additional validation &
Limited; attention weights alone are insufficient for verification or assurance arguments &
Relevant for transformer-based detectors such as DETR, but should be treated as diagnostic signals rather than faithful explanations \\

\hline

Surrogate models, e.g., LIME, SHAP &
Medium to high; provide simplified local or feature-level explanations &
Medium to low; dependent on sampling strategy, neighborhood definition, and approximation stability &
Medium; local approximations can be manipulated or fail outside the sampled neighborhood &
Moderate; useful for structured analysis if approximation fidelity is explicitly validated &
Potentially useful for feature-level analysis and metadata-supported ATR, but challenging in high-dimensional image, radar, or multisensor settings \\

\hline

Detection-aware XAI, e.g., detector-specific CAM variants, G-CAME, OD-specific evaluation methods &
High for object localization; explanations can be tied to bounding boxes or detections &
Medium; better aligned with object detection outputs, but still affected by perturbations and model sensitivity &
Medium to high; localization-specific explanations can still be visually plausible but causally weak &
Moderate; more suitable than generic saliency for detection tasks, but still requires robustness and faithfulness tests &
Strong fit for ATR object detection when spatial localization is operationally relevant \\

\hline

Intrinsic/ physics-informed models &
Medium to high; explanations are embedded in model structure or physical assumptions &
High if physical constraints match the sensing process &
Low to medium; harder to manipulate if explanations are constrained by domain knowledge and signal formation &
High; better suited for assurance, traceability, and systematic validation &
Particularly suitable for radar, SAR, multisensor ATR, and safety-critical applications where physical consistency is required \\

\hline
\end{tabular}
\end{table}

\subsection{Summary Table}

To provide a structured overview, Table \ref{tab:xai_assessment_atr} summarizes the considered XAI para\-digms with respect to the evaluation criteria introduced in Section \ref{sec: Evaluation Criteria}.

Table \ref{tab:xai_assessment_atr} reveals a clear trade-off between accessibility and reliability. Methods that are easy to apply and visually intuitive, such as saliency- and attention-based approaches, tend to fall short in terms of robustness, causal validity, and verifiability. Detection-aware XAI methods, including detector-specific CAM variants and approaches such as G-CAME, improve spatial alignment with object detections and enhance interpretability in detection tasks, but do not fully resolve these fundamental limitations. Surrogate models increase transparency through simplified approximations, yet remain dependent on approximation fidelity and local sampling assumptions. In contrast, intrinsic and physics-informed approaches are better aligned with the requirements of safety-critical ATR, as they ground explanations in domain knowledge and signal formation processes.

Overall, this comparison demonstrates the need for a shift from visually convincing explanations toward methods that provide robust, causally meaningful, and operationally relevant insight.

\section{Discussion and Failure Modes}

The analysis above indicates that current XAI approaches exhibit systematic weaknesses when applied to safety-critical ATR systems. Beyond quantitative evaluation criteria, it is essential to consider typical failure modes that arise in practice and may compromise the reliability and interpretability of explanations.

\subsection{Spurious Explanations}

A central issue of many post-hoc explanation methods is the occurrence of spurious explanations. These are explanations that appear visually plausible but do not accurately reflect the underlying decision process of the model. In particular, gradient-based saliency methods may produce structured patterns that are largely influenced by the input distribution rather than the learned model parameters.

Previous work has demonstrated that some saliency methods can generate similar explanations even when model weights are randomized \cite{adebayo2018sanity}, indicating a weak dependence on the actual model behavior. In the ATR context, this can lead to misleading interpretations, where explanations highlight features that are not causally relevant for target recognition. Such discrepancies pose a significant risk in safety-critical applications, where decisions must be both accurate and interpretable.

\subsection{Overtrust in Visual Explanations}

Another important failure mode is overtrust induced by visually convincing explanations. Heatmaps and attention visualizations are often intuitive and easy to interpret, which may lead users to overestimate their reliability.

Empirical studies suggest that humans tend to trust explanations that are simple and visually coherent, even when they are incomplete or incorrect. In ATR systems, this can result in operators placing unjustified confidence in model decisions, potentially leading to incorrect or unsafe actions. The combination of high visual plausibility and limited causal validity therefore creates a systematic risk of misinterpretation.

\subsection{Instability and Sensitivity to Perturbations}

Many XAI methods exhibit significant sensitivity to small input perturbations. Minor changes in the input data, such as noise, slight shifts, or variations in viewing conditions, can lead to substantially different explanations.

This instability is particularly problematic in ATR applications, where sensor data is inherently noisy and subject to environmental variability. Explanations that are not robust under such conditions cannot be reliably used for system validation or decision support. Moreover, instability reduces the reproducibility of explanations, which is a critical requirement in safety-critical systems.

\subsection{Limits of Explainability in Deep Learning Systems}

Beyond method-specific limitations, there exist fundamental constraints on explainability in complex deep learning models. Modern neural networks operate in high-dimensional and highly nonlinear representation spaces, where decision boundaries cannot always be decomposed into simple, human-interpretable components.

Furthermore, multiple internal representations may lead to equivalent model outputs, making it difficult to establish unique causal explanations. In many cases, no ground truth for explanations exists, which complicates both evaluation and validation. These limitations suggest that post-hoc explanation methods may inherently fall short in fully capturing model behavior in complex ATR systems.

\subsection{Implications for Safety-Critical ATR Systems}

The identified failure modes have direct implications for the deployment of XAI in safety-critical ATR applications. In such systems, explanations must not only be interpretable but also reliable, stable, and aligned with the actual decision process of the model.

The analysis indicates that relying solely on post-hoc explanation methods is insufficient for high-assurance applications \cite{rudin2019stop}. Instead, there is a need for approaches that integrate interpretability into the model design, incorporate domain knowledge, and support formal validation and verification processes.

Overall, these findings reinforce the need for a shift toward more robust, causally grounded, and physically informed explainability methods, particularly in domains where incorrect interpretations may have severe consequences.

\section{Conclusion and Outlook}

This work examined the applicability of current XAI methods in safety-critical ATR systems. We introduced a taxonomy of commonly used approaches, including saliency-based, attention-based, surrogate-based, detection-aware, and intrinsic or physics-informed methods, and proposed an ATR-specific assessment framework with respect to interpretability, robustness, vulnerability to manipulation, and suitability for validation and verification.

The analysis revealed fundamental limitations of current post-hoc explainability methods. While saliency- and attention-based approaches provide intuitive and easily deployable explanations, they often lack robustness, causal validity, and reliability in safety-critical scenarios. Detection-aware XAI methods improve alignment with object detection outputs and enhance spatial interpretability, but still inherit key limitations of underlying gradient-based techniques, including sensitivity to perturbations and limited causal grounding. Surrogate models increase transparency but remain dependent on approximation fidelity, which can be problematic in complex, high-dimensional ATR settings. In contrast, intrinsic and physics-informed approaches show greater potential for reliable and consistent explanations, particularly when aligned with domain knowledge.

The discussion of failure modes further highlighted critical challenges, including spurious explanations, overtrust in visually plausible outputs, and instability under perturbations. These findings indicate that explainability in ATR systems must be evaluated not only in terms of interpretability, but also with respect to robustness and operational reliability.

Future research should therefore focus on the development of hybrid and intrinsically interpretable models that combine the flexibility of deep learning with structured, physically grounded representations. In addition, there is a need for standardized evaluation frameworks and benchmarks tailored to safety-critical applications, enabling systematic comparison and validation of XAI methods, including detection-aware approaches.

Advancing XAI in ATR will require moving beyond perceptually plausible explanations toward explanations that are robust, causally meaningful, verifiable, and grounded in the underlying physical processes of the data.

\bibliographystyle{splncs04}
\bibliography{references}

\end{document}